\documentclass{article}

\usepackage[preprint]{conference}

\usepackage{microtype}
\usepackage{hyperref}
\usepackage{url}
\usepackage{booktabs}
\usepackage{multirow}

\usepackage{amsmath}
\usepackage{amsthm}
\usepackage{amssymb}
\usepackage{amsfonts}
\usepackage{dsfont}
\usepackage{graphicx}
\usepackage{adjustbox}
\usepackage{caption}
\usepackage{subcaption}

\definecolor{darkblue}{rgb}{0, 0, 0.5}
\hypersetup{colorlinks=true, citecolor=darkblue, linkcolor=darkblue, urlcolor=darkblue}
\usepackage{dsfont}

\theoremstyle{plain}
\newtheorem{theorem}{Theorem}

\newtheorem{proposition}[theorem]{Proposition}
\newtheorem*{proposition*}{Proposition}

\theoremstyle{definition}

\newtheorem*{definition*}{Definition}

\theoremstyle{remark}

\title{Cost-Saving LLM Cascades with Early Abstention}

\author{Michael J. Zellinger \& Rex Liu\\
Computing + Mathematical Sciences\\
California Institute of Technology\\
\texttt{\{zellinger, rex\}@caltech.edu} \\
\And
Matt Thomson \\
Biology \& Biological Engineering \\
California Institute of Technology \\
\texttt{mthomson@caltech.edu} \\
}

\date{}

\RequirePackage{natbib}
\begin{document}

\ifcolmsubmission
\linenumbers
\fi

\maketitle

\begin{abstract}
LLM cascades deploy small LLMs to answer most queries, limiting the use of large and expensive LLMs to difficult queries. This approach can significantly reduce costs without impacting performance. However, risk-sensitive domains such as finance or medicine place an additional premium on avoiding model errors. Since even the most expensive models are susceptible to making mistakes, applications in these domains benefit from allowing LLM systems to completely abstain from answering difficult queries. Introducing abstention poses a design question for LLM cascades: should abstention only be allowed at the final model or also at earlier models? Since error patterns of small and large models are correlated, allowing earlier models to abstain may reduce inference costs and latency by anticipating abstention decisions by expensive and slow models-- avoiding the need to run these models. We investigate the benefits of such ``early abstention'' in LLM cascades and find that it reduces overall test loss by 2.2\% on average across six benchmarks (GSM8K, MedMCQA, MMLU, TriviaQA, TruthfulQA, and XSum). These gains result from a more effective use of abstention, trading a 4.1\% average increase in the overall abstention rate for a 13.0\% reduction in cost and a 5.0\% reduction in error rate. Our findings demonstrate the possibility of leveraging correlations between the error patterns of different language models to drive performance improvements for LLM systems with abstention.
\end{abstract}

\maketitle

\section{Introduction}

Following the proliferation of publicly available LLMs, many authors have proposed combining different models into LLM systems to enhance reliability and reduce costs. Specifically, LLM cascades have emerged as a powerful design pattern for navigating the trade-off between performance and cost when deploying LLMs (\citealp{chen2023}; \citealp{madaan2024}; \citealp{yue2024}; \citealp{jitkrittum2024}), especially for edge-to-cloud deployments (\citealp{kag2023}).

The basic idea behind cascades is that it would be inefficient to process all queries using the largest and most expensive LLMs. On many tasks, relatively small LLMs can correctly answer most queries. To leverage this fact, a cascade initially sends each query to a small LLM. If this small model has a high chance of correctly answering the query, as measured by a confidence indicator, the cascade returns the output to the user. Otherwise, the cascade defers the query to a larger and more expensive model. In this way, LLM cascades yield impressive cost and latency reductions without meaningfully compromising performance (\citealp{chen2023}). However, cost and latency reduction are not the only desiderata of an LLM cascade. In risk-sensitive areas such as finance, law, or medicine, the overriding concern is often to avoid mistakes. For these applications, it is natural to give LLM cascades the ability to abstain from answering any query when the risk of making a mistake is significant (\citealp{chow1957}, \citealp{chow1970}, \citealp{elyaniv2010}, \citealp{cortes2016}).

Introducing abstention poses a design question for LLM cascades: is it sufficient to allow the final model to abstain, or should the smaller models at the start of the cascade be allowed to abstain ``early''? Since error patterns of small and large models are correlated (\citealp{zellinger2025}), there is a potential for the small models to anticipate abstention decisions by large models, significantly reducing inference cost and latency on abstained queries.

Understanding the benefits of such \textit{early abstention} is important for cascade design since implementing early abstention requires access to a cascade's internal logic-- it cannot simply be added on later to a black-box cascade. By contrast, abstaining at the final model can be implemented as a simple post-processing step of the final model's output and its associated confidence score. Hence, an understanding of the benefits of early abstention helps inform whether the capacity for abstention can be effectively \textit{decoupled} from the overall implementation of a cascade.

In this paper, our main contributions are as follows:
\begin{itemize}
    \item We introduce the possibility of ``early abstention'' for LLM cascades with abstention.
    \item We extend the continuous optimization-based algorithm for tuning confidence thresholds in LLM cascades of \cite{zellinger2025} to a multi-objective formalism incorporating abstention.
    \item We demonstrate that early abstention lowers the overall test loss by 2.2\% across six benchmarks (GSM8K, MedMCQA, MMLU, TriviaQA, TruthfulQA, XSum).
    \item We show that early abstention allows cascades to make more effective use of the ability to abstain, trading an increase in the abstention rate (+4.1\% on average) for significant cost reductions (-13.0\% on average) and error reductions (-5.0\% on average).
\end{itemize}

\section{Background and Related Work}

This paper combines two distinct strands of research within the language modeling community: selective prediction, which aims at reducing the error rate by allowing models to abstain on the most difficult queries, and LLM cascades, which focus on combining small and large LLMs to reduce cost.

\noindent \textbf{Selective prediction.} Selective prediction focuses on giving machine learning models the ability to abstain on difficult queries (\cite{elyaniv2010}). This perspective is especially useful in risk-sensitive domains, where the cost of making a mistake far outweighs the inconvenience of handling abstained queries. The field goes back to the works of Chow (\cite{chow1957}, \cite{chow1970}), who analyzed abstention in the context of applying optical character recognition to scanning business documents. \cite{geifman2017} applied selective prediction in deep learning, significantly reducing image classification error rates.

In natural language processing, \cite{xin2021} and \cite{yoshikawa2023} introduced abstention for language models by thresholding various confidence scores derived from the conditional token distribution. In the modern era of \textit{large} language models post-ChatGPT, selective prediction research has largely taken place in the emerging field of uncertainty quantification for LLMs (\citealp{manakul2023}; \citealp{farquhar2024}; \citealp{lin2024}).

\noindent \textbf{LLM cascades.} LLM cascades leverage confidence scores in a different way from selective prediction. Rather than abstain on difficult queries, a cascade adaptively selects the most cost-effective model to answer each query. An incoming query first goes to a small LLM. If the small model's confidence score is below a chosen threshold, the cascade forwards the query to a larger and more powerful LLM. This approach has yielded impressive cost savings without impacting performance (\citealp{chen2023}; \cite{madaan2024}; \citealp{yue2024}). Cascades often make use of an LLM's conditional probability distribution for computing confidence scores (\citealp{jitkrittum2024}), finetune smaller language models for correctness prediction (\citealp{chen2023}), or compute consistency-based uncertainty metrics (\citealp{yue2024}).

However, few prior works raise the question of correlations between the confidence scores of different LLMs, with the exception of \cite{zellinger2025}. Our approach benefits from such correlations, as they make it easier for small models to anticipate abstention decisions by larger models.

\section{Cost-Saving LLM Cascades with Abstention}
\label{sec:methodology-overview}

\noindent \textbf{Cascades with abstention.} Let $C = M_1 \rightarrow ... \rightarrow M_k$ be a large language model (LLM) cascade. Given a query $x$, each model $M_i$ uses a confidence score $\Phi_i = \Phi_i(x) \in [0,1]$ to decide whether to return the response or defer the query to $M_{i+1}$. In a typical LLM cascade without abstention, $M_i$ returns the response if its confidence $\Phi_i$ exceeds its deferral threshold $\phi_i$. 

To introduce the possibility of abstention, we pair each deferral threshold $\phi_i$ with an additional \textit{abstention threshold} $\xi_i < \phi_i$. The intuition is that if confidence is \textit{low}, the model should defer, but if it is \textit{very low}, the model should abstain. Given a query $x$, the output $C(x)$ of a cascade $C = M_1 \rightarrow ... \rightarrow M_k$ with deferral thresholds $(\phi_1, ..., \phi_{k-1})$ and abstention thresholds $(\xi_1, ..., \xi_{k})$ is then defined recursively as
\begin{equation}
    \label{eq:cascade_model}
    C(x) = \begin{cases}
        M_1(x) & \text{~if~} \Phi_1(x) > \phi_1 \text{~or~} \vert C \vert = 1\\
        C_{\text{2:k}}(x) & \text{~if~} \xi_1 \leq \Phi_1(x) \leq \phi_1, \\
        \varnothing & \text{~if~} \Phi_1(x) < \xi_1,
    \end{cases}
\end{equation}
where $C_\text{2:k}$ is the subcascade $M_2 \rightarrow ... \rightarrow M_k$, and returning $\varnothing$ means that the cascade abstains. 

In other words, the cascade abstains if the confidence score for any model $M_i$ falls below its abstention threshold $\xi_i$. If the confidence score exceeds $\xi_i$ but is still below the deferral threshold $\phi_i$, the query is forwarded to $M_{i+1}$. Note that the final model $M_k$ of the cascade has an abstention threshold $\xi_k$ but no deferral threshold, since there is no downstream model to which it could defer the query.

\begin{definition*}[Early Abstention]
Given a cascade $C = M_1 \rightarrow ... \rightarrow M_k$ with deferral thresholds $\boldsymbol{\phi} = (\phi_1, ..., \phi_{k-1})$ and abstention thresholds $\boldsymbol{\xi} = (\xi_1, ..., \xi_{k})$, we define \textit{early abstention} as the event
\begin{equation}
    \mathcal{A}_E = \bigcup_{i=1}^{k-1}\{ x : \Phi_i(x) < \xi_i\}
\end{equation}
that the confidence $\Phi_i$ of an earlier model $(i < k)$ falls below its abstention threshold $\xi_i$.
\end{definition*}

Compared to abstaining only at the final model, early abstention can reduce the cost and latency of a cascade, since confidence scores between small and large LLMs are often correlated (\citealp{zellinger2025}). In Section \ref{subsec:inefficiency-of-final-model-abstention}, we present empirical precision-recall trade-offs for the ability to anticipate a large model's abstention decisions.

\noindent \textbf{Evaluating cascade performance.} Introducing abstention adds complexity to performance evaluation, as the overall performance of a cascade now depends on the error rate, cost, and abstention rate-- a three-dimensional space. Consistent with standard multi-objective optimization, given two performance vectors $\nu=(e, c, a)$ and $\nu'=(e', c', a')$ we say that $\nu$ \textit{dominates} $\nu'$ if $e \leq e'$, $c \leq c'$, and $a \leq a'$, and at least one of the inequalities is strict. Any performance vector $\nu$ that is not dominated by any other performance vector is \textit{Pareto-optimal}.

A cascade $C$ gives rise to a set of Pareto-optimal performance vectors, corresponding to the possible choices for deferral and abstention thresholds. For example, raising deferral thresholds generally results in lower error rates while increasing cost; raising abstention thresholds leads to lower error rates while increasing the abstention rate. The set of Pareto-optimal performance vectors 
\begin{equation}
    \Gamma(C) = \{ \nu \in \mathbb{R}^3: \nu \text{~is Pareto-optimal for $C$} \}
\end{equation}
forms a nonlinear two-dimensional surface embedded in $\mathbb{R}^3$. These surfaces can be cumbersome to compare across different cascades. We provide a more user-friendly approach by introducing the user preference parameters $\lambda_\text{c} \geq 0$ and $\lambda_\text{a} \geq 0$ measuring the user's sensitivity towards higher cost and query abstention, respectively, relative to the need for avoiding errors.

\begin{definition*}[Cascade Loss]
We measure performance of a cascade $C$, given user preferences $\lambda_{c}$, $\lambda_{a} \geq 0$ for avoiding cost and query abstentions, by the \textit{cascade loss}
    \begin{equation}
    \label{eq:cascade_loss}
    \ell(\lambda_{c}, \lambda_{a}) = \min_{\boldsymbol{\phi}, \boldsymbol{\xi}} \mathbb{P}(\text{Error} \land \neg \text{Abstention}) + \lambda_\text{c}~ \mathbb{E}[\text{Cost}] + \lambda_\text{a}~ \mathbb{P}(\text{Abstention}),
\end{equation}
where $[\boldsymbol{\phi}, \boldsymbol{\xi}] = [\phi_1, ..., \phi_{k-1}, \xi_1, ..., \xi_{k}] \in \mathbb{R}^{2k-1}$ are the deferral and abstention thresholds. Given a grid of user preferences $\Lambda_{c} \times \Lambda_{a}$, we measure overall system performance by the \textit{overall cascade loss}
\begin{equation}
    \label{eq:overall_cascade_loss}
    \mathcal{L}(C) = \frac{1}{\vert \Lambda_{c} \vert \vert \Lambda_{a} \vert } \sum_{(\lambda_{c}, \lambda_{a}) \in \Lambda_c \times \Lambda_a} \ell(\lambda_{c}, \lambda_{a}),
\end{equation}
\end{definition*}
which averages the cascade loss over the grid of user preferences.

\begin{figure}
    \centering
    \includegraphics[width=0.8\textwidth]{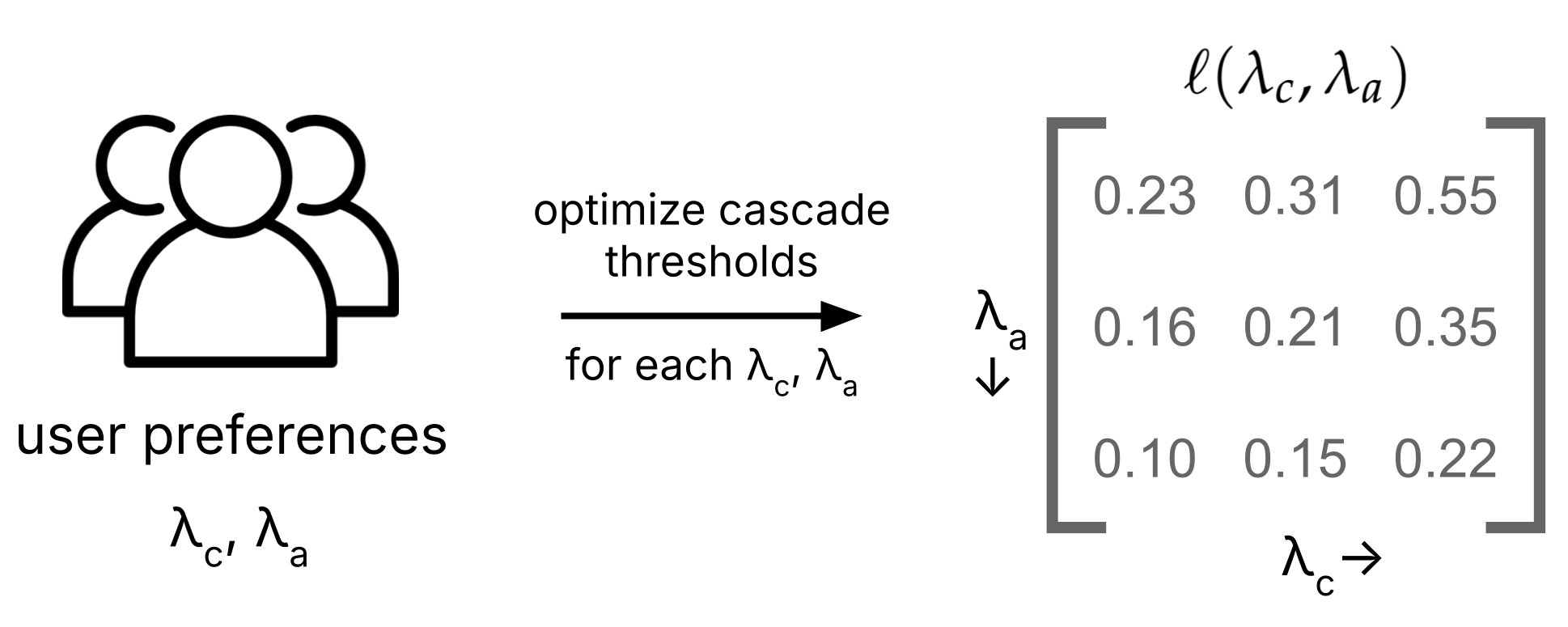}
    \caption{Our framework evaluates cascade performance across the range of user preferences by taking into account the user's desire to reduce cost ($\lambda_{c}$) and abstention ($\lambda_{a}$), relative to the need for avoiding errors.}
    \label{fig:cascade_loss}
\end{figure}

Equation (\ref{eq:cascade_loss}) aligns with previous work on selective prediction (\citealp{cortes2016}), while the overall cascade loss (\ref{eq:overall_cascade_loss}) provides an effective means for comparing different cascades, while avoiding the difficulties of nonlinear Pareto surfaces. Figure \ref{fig:cascade_loss} illustrates our approach for evaluating cascade performance across the user preference space. Importantly, each pair of preferences $(\lambda_{c}, \lambda_{a})$ requires separately optimizing the cascade thresholds $[\boldsymbol{\phi}, \boldsymbol{\xi}] \in \mathbb{R}^{2k-1}$.

\noindent \textbf{Optimizing cascade thresholds}. Computing the cascade loss $\ell(\lambda_c, \lambda_a)$ requires selecting deferral and abstention thresholds $[\boldsymbol{\phi}, \boldsymbol{\xi}] \in \mathbb{R}^{2k-1}$ for each choice of user preferences $(\lambda_c, \lambda_a)$. Intuitively, penalizing cost ($\lambda_{c} \uparrow$) results in lower deferral thresholds (increased use of smaller models), whereas penalizing abstention ($\lambda_{a} \uparrow$) leads to lower abstention thresholds (less abstention). To optimize $[\boldsymbol{\phi}, \boldsymbol{\xi}]$ for each $(\lambda_c, \lambda_a)$, we apply the continuous optimization-based algorithm of \cite{zellinger2025}. Their approach is based on 

\begin{enumerate}
    \item Modeling the joint distribution of confidence scores $(\Phi_1, ..., \Phi_k)$ through a Markov factorization based on mixture-of-beta marginals $p(\Phi_i)$ and copula-based conditionals $p(\Phi_i | \Phi_{i-1})$:
    \begin{equation*}
        p(\Phi_1, ..., \Phi_k) \approx p(\Phi_1) \prod_{i=2}^{k} p(\Phi_i | \Phi_{i-1}).
    \end{equation*}
    \item Plugging $[\boldsymbol{\phi}$,$\boldsymbol{\xi}$]-differentiable expressions for the error probability, expected cost, and abstention probability into the cascade loss (\ref{eq:cascade_loss}) and finding the optimal cascade thresholds $\boldsymbol{\phi}^{*}, \boldsymbol{\xi}^{*}$ using a continuous optimizer.
\end{enumerate}

We apply this continuous optimization-based approach largely because a simple grid search faces runtime issues with increasing numbers of parameters. Implementing early abstention essentially doubles the number of cascade thresholds from $k-1$ (deferral thresholds only) to $2k-1$ (deferral and abstention thresholds), so the cross-over point when grid search becomes inconveniently slow happens much sooner (as early as $k=3$).

Note that \cite{zellinger2025} do not consider abstention, so we extend their methodology to include abstention thresholds. We provide the modified Markov equations for the the correctness probability, expected cost, and abstention rate in Appendix A. Solving the optimization problem (\ref{eq:cascade_loss}) requires enforcing the inequalities $\xi_i < \phi_i$ for $i<k$ (each abstention threshold must be lower than its corresponding deferral threshold). To implement these constraints, we switch the optimizer to sequential least squares (\citealp{scipy2020}).

\noindent \textbf{Quantifying confidence}. To compute LLM confidence scores, we follow the approach of \cite{zellinger2025}, who use logistic regression to calibrate a token-based confidence signal $p_\text{raw}(x)$, which varies based on the type of task:
\begin{itemize}
    \item \textbf{multiple-choice question answering}: $p_\text{raw}$ is the maximum probability assigned to the valid answer tokens (for example, ``A'', ``B'', ``C'', or ``D'' for MMLU)
    \item \textbf{free-form natural language generation}: the LLM receives a follow-up verification query to assess whether its proposed response is correct; $p_\text{raw}$ is the token probability assigned to ``Y'' (yes, correct) as opposed to ``N'' (no, incorrect)
\end{itemize}

In principle, any other confidence signal could be used, such as the consistency-based confidence measures proposed by \cite{manakul2023} and \cite{farquhar2024}.

\section{Results}
\label{sec:results}

First, we demonstrate the inefficiency of abstaining at the final model by showing that smaller models are capable of anticipating a large model's decision to abstain. Second, we compare the overall cascade loss between cascades \textit{with} and \textit{without} early abstention, finding that
\begin{itemize}
    \item Early abstention improves overall cascade performance by 2.2\% on average, ranging from a 2\% detriment to a 7.3\% gain across six benchmarks.
    \item The benefits of early abstention are visibly concentrated in the upper right quadrant of the user preference space ($\lambda_{c}$ high, $\lambda_{a}$ low), where cost is a concern and abstention is not heavily penalized.
    \item Early abstention makes more effective use of the ability to abstain, trading a 4.1\% increase in the overall abstention rate for an average cost reduction of 13.0\% and an average error reduction of 5.0\%.
\end{itemize}
To carry out our experiments, we build on the data from \cite{zellinger2025}. All results are based on a low-data regime with $n\approx300$ training examples, and LLMs are prompted zero-shot.

\subsection{The Inefficiency of Final-Model Abstention}
\label{subsec:inefficiency-of-final-model-abstention}

\begin{figure*}[t]
    \centering
    \begin{subfigure}{0.3\textwidth}
        \centering
        \includegraphics[width=\linewidth]{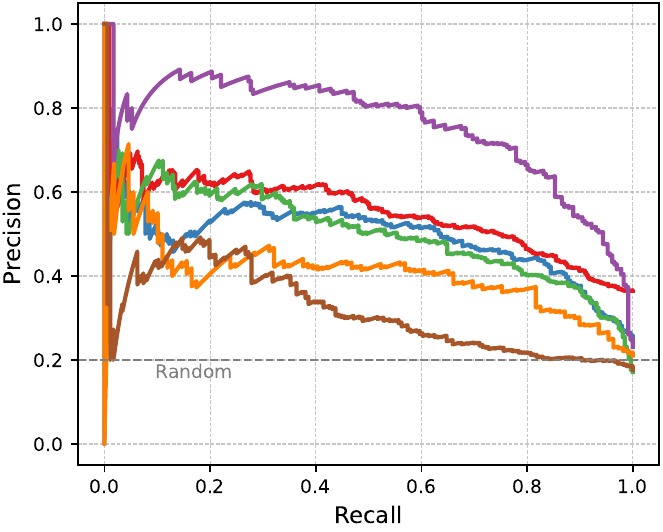}
        \subcaption{Llama3.1 405B, 20\%}
    \end{subfigure}
    \hfill
    \begin{subfigure}{0.3\textwidth}
        \centering
        \includegraphics[width=\linewidth]{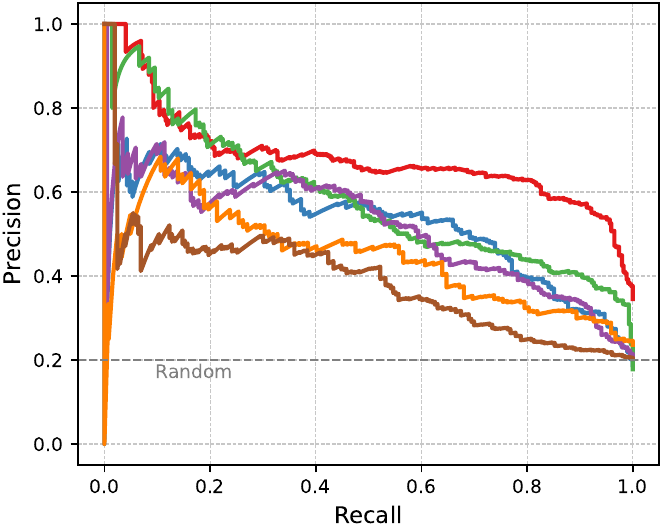}
        \subcaption{OpenAI GPT-4o, 20\%}
    \end{subfigure}
    \hfill
    \begin{subfigure}{0.3\textwidth}
        \centering
        \includegraphics[width=\linewidth]{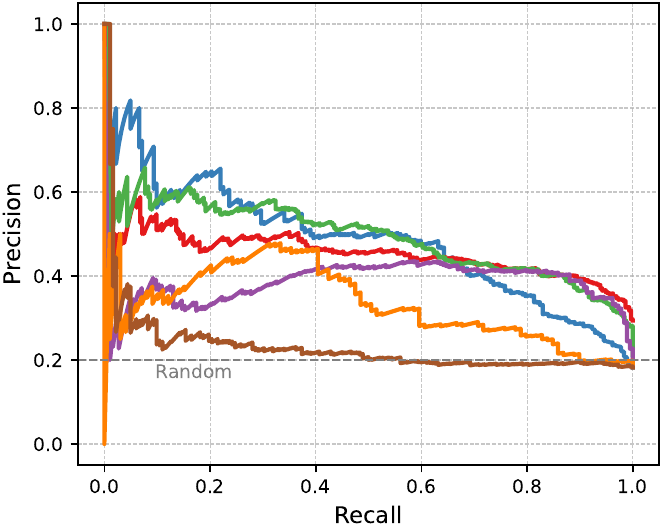}
        \subcaption{Qwen2.5 72B, 20\%}
    \end{subfigure}

    \vspace{1em}
    
    \begin{subfigure}{0.3\textwidth}
        \centering
        \includegraphics[width=\linewidth]{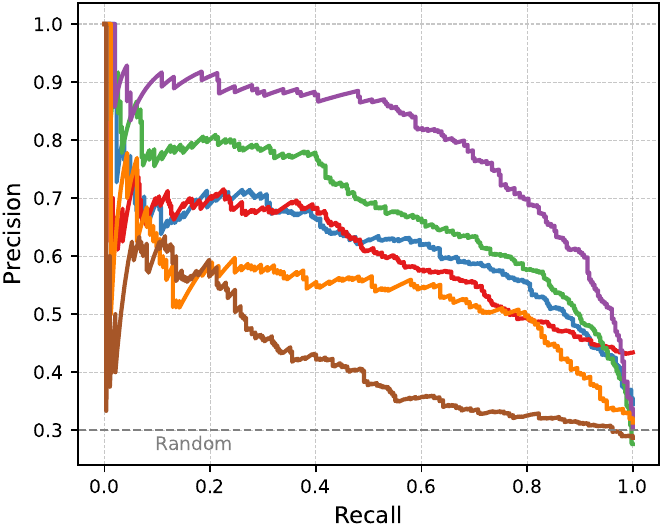}
        \subcaption{Llama3.1 405B, 30\%}
    \end{subfigure}
    \hfill
    \begin{subfigure}{0.3\textwidth}
        \centering
        \includegraphics[width=\linewidth]{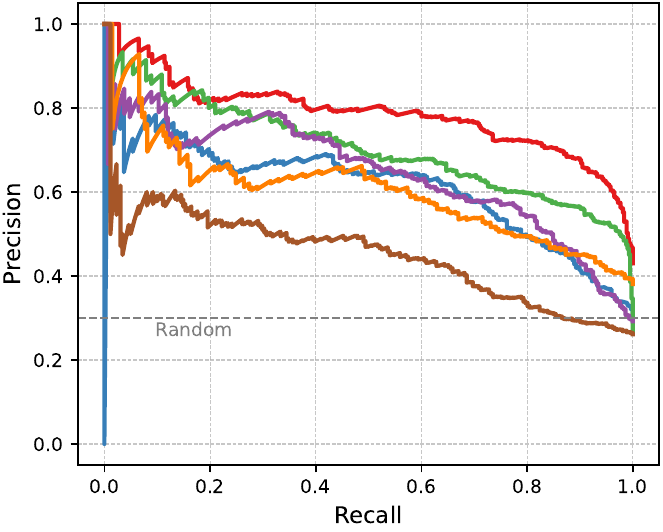}
        \subcaption{OpenAI GPT-4o, 30\%}
    \end{subfigure}
    \hfill
    \begin{subfigure}{0.3\textwidth}
        \centering
        \includegraphics[width=\linewidth]{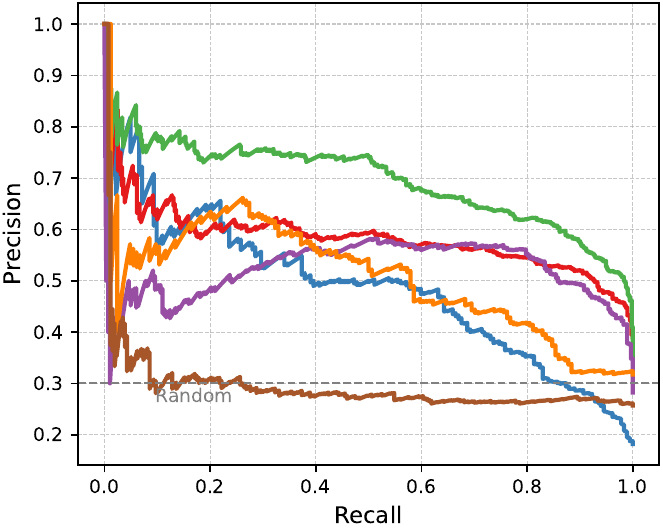}
        \subcaption{Qwen2.5 72B, 30\%}
    \end{subfigure}
    
    \vspace{1em}
    
    \includegraphics[width=1.0\textwidth]{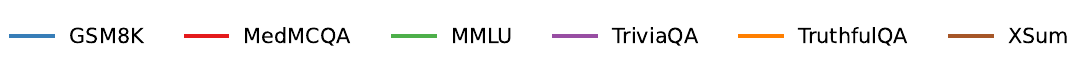}
    
    \caption{Precision-recall curves for \textit{early abstention}-- predicting a large model's decision to abstain using small models' confidence scores; \% is the large model's abstention rate. Performance visibly exceeds the random baseline (dashed gray line, equal to large model's abstention rate). At lower recall, smaller models can often predict final-model abstentions with high precision.}
    \label{fig:pr-cascades}
\end{figure*}

To assess the ability of small models to anticipate abstention decisions by a larger model, we compute empirical precision-recall trade-offs for abstention prediction for the following LLM cascades:
\begin{itemize}
    \item Meta Llama3 1B $\rightarrow$ 3B $\rightarrow$ 8B $\rightarrow$ 70B $\rightarrow$ \textbf{405B}
    \item OpenAI GPT-4o Mini $\rightarrow$ \textbf{GPT-4o}
    \item Alibaba Qwen2.5 32B Coder $\rightarrow$ \textbf{Qwen 72B}
\end{itemize}

In our experiment, we set the abstention threshold of the final model (highlighted in bold above) so that it abstains on the bottom 20\% and 30\% of queries with the lowest confidence scores on the training set. We then train logistic regression classifiers to predict these abstention decisions, based on the confidence scores of all previous models in the cascade. The training sets contain $n\approx300$ examples and the test sets $n\approx1000$ examples. See Appendix B for implementation details.

Figure \ref{fig:pr-cascades} shows the precision-recall trade-offs computed on the test sets. We evaluate abstention prediction on six benchmarks (GSM8K, MMLU, MedMCQA, TriviaQA, TruthfulQA, and XSum) spanning a variety of tasks from mathematical reasoning to summarization. 

The results suggest that at low recall, smaller models can often predict final-model abstention with high precision. To see the significance of these results, consider the case of 80\% precision at 20\% recall. Early abstention leads to cost savings on the correct calls ($30\% \text{~abstention rate} \times 20\% \text{~recall}=6\%$ of all queries) and incorrect calls ($25\% \text{~imprecision/precision} \times 6\% \text{~correct calls}= 1.5\% \text{~of all queries}$). Each instance of early abstention reduces inference costs by approximately a factor of ten. Hence, early abstention reduces overall inference costs roughly from $C$ to $0.925 \times C + 0.075 \times (C/10)$: a $6.75\%$ total cost reduction in exchange for raising the overall abstention rate from 30\% to 31.5\%.

\subsection{Evaluating the Benefits of Early Abstention}

To formally evaluate the benefits of early abstention, we compare the performance of two different architectures for LLM cascades $M_1 \rightarrow ... \rightarrow M_k$:
\begin{itemize}
    \item \textbf{Early Abstention (``Early'')}: each model $M_i$ has a separate abstention threshold $\xi_i$
    \item \textbf{Final-Model Abstention (``Final'')}: only $M_k$ has an abstention threshold $\xi_k$
\end{itemize}
To compare cascade performance, we optimize the cascade thresholds across a grid of user preference parameters $\lambda_c, \lambda_a$ as described in Section \ref{sec:methodology-overview}. We applied light smoothing on the 2D grid of optimal thresholds by detecting outliers\footnote{We call a threshold vector an \textit{outlier} if the mean of its components differs from the means of neighboring threshold vectors by much more than the variability of those neighbors. More details can be found in Appendix C.} and replacing their values with the mean values of their neighbors. To assess the overall performance of a cascade, we compute the overall cascade loss (\ref{eq:overall_cascade_loss}). All reported numbers refer to performance on the the test sets ($n \approx 1000$ examples). We optimized the cascade thresholds on training sets with $n \approx 300$ examples. 

\noindent \textbf{Model names.} In our tables, \textit{LxB} are the Meta Llama3 models with $x$ billion parameters; \textit{4o} and \textit{4o Mini} are the OpenAI GPT models; and \textit{QxB(c)} are the Alibaba Qwen2.5 (coding) models with $x$ billion parameters.

\begin{table}[t]
\centering
\small
\caption{Change in overall cascade loss on the test set when using early abstention (``Early'') 
compared to final-model abstention (``Final''): GSM8K, MedMCQA, and MMLU.}
\label{tab:overall-loss-1}
\begin{adjustbox}{max width=\textwidth}
\begin{tabular}{lrrrrrrrrrr}
\toprule
\multirow{2}{*}{\textbf{Cascade}} 
 & \multicolumn{3}{c}{\textbf{GSM8K}} 
 & \multicolumn{3}{c}{\textbf{MedMCQA}} 
 & \multicolumn{3}{c}{\textbf{MMLU}} \\
 & \textbf{Early} & \textbf{Final} & \textbf{\%$\boldsymbol{\Delta}$}
 & \textbf{Early} & \textbf{Final} & \textbf{\%$\boldsymbol{\Delta}$}
 & \textbf{Early} & \textbf{Final} & \textbf{\%$\boldsymbol{\Delta}$} \\
\midrule
\textbf{L1B $\rightarrow$ L3B} & {\textbf{0.196}} & {0.199} & {-1.772}
                  & {0.349} & {\textbf{0.349}} & {0.054}
                  & {0.319} & {\textbf{0.319}} & {0.066} \\
\textbf{L1B $\rightarrow$ L8B} & {\textbf{0.157}} & {0.158} & {-0.189}
                   & {0.355} & {\textbf{0.354}} & {0.242}
                   & {0.279} & {\textbf{0.278}} & {0.243} \\
\textbf{L1B $\rightarrow$ L70B} & {\textbf{0.108}} & {0.111} & {-2.476}
                    & {\textbf{0.243}} & {0.244} & {-0.441}
                    & {\textbf{0.177}} & {0.178} & {-0.614} \\
\textbf{L1B $\rightarrow$ L405B} & {\textbf{0.186}} & {0.211} & {-12.057}
                     & {\textbf{0.271}} & {0.285} & {-5.042}
                     & {\textbf{0.199}} & {0.210} & {-5.243} \\
\textbf{L3B $\rightarrow$ L8B} & {0.182} & {\textbf{0.182}} & {0.350}
                   & {\textbf{0.355}} & {0.355} & {-0.065}
                   & {0.276} & {\textbf{0.275}} & {0.300} \\
\textbf{L3B $\rightarrow$ L70B} & {0.120} & {\textbf{0.115}} & {4.369}
                    & {0.247} & {\textbf{0.245}} & {1.064}
                    & {\textbf{0.174}} & {0.177} & {-1.520} \\
\textbf{L3B $\rightarrow$ L405B} & {\textbf{0.161}} & {0.170} & {-5.615}
                     & {\textbf{0.270}} & {0.286} & {-5.662}
                     & {\textbf{0.191}} & {0.201} & {-4.904} \\
\textbf{L8B $\rightarrow$ L70B} & {\textbf{0.111}} & {0.111} & {-0.090}
                    & {0.254} & {\textbf{0.252}} & {0.502}
                    & {\textbf{0.178}} & {0.182} & {-2.292} \\
\textbf{L8B $\rightarrow$ L405B} & {\textbf{0.131}} & {0.136} & {-3.735}
                     & {\textbf{0.281}} & {0.298} & {-5.606}
                     & {\textbf{0.185}} & {0.197} & {-5.982} \\
\textbf{L70B $\rightarrow$ L405B} & {0.103} & {\textbf{0.102}} & {0.720}
                      & {\textbf{0.260}} & {0.269} & {-3.682}
                      & {\textbf{0.169}} & {0.184} & {-8.250} \\
\textbf{4o-Mini $\rightarrow$ Q32Bc} & {\textbf{0.093}} & {0.096} & {-3.895}
                         & {\textbf{0.282}} & {0.286} & {-1.433}
                         & {\textbf{0.207}} & {0.209} & {-1.090} \\
\textbf{4o-Mini $\rightarrow$ Q72B} & {0.091} & {\textbf{0.088}} & {3.691}
                        & {\textbf{0.259}} & {0.260} & {-0.410}
                        & {\textbf{0.169}} & {0.169} & {-0.114} \\
\textbf{4o-Mini $\rightarrow$ 4o} & {\textbf{0.093}} & {0.097} & {-4.833}
                       & {\textbf{0.225}} & {0.231} & {-2.677}
                       & {\textbf{0.174}} & {0.180} & {-3.673} \\
\textbf{Q32Bc $\rightarrow$ Q72B} & {0.116} & {\textbf{0.103}} & {11.807}
                       & {\textbf{0.287}} & {0.288} & {-0.207}
                       & {\textbf{0.193}} & {0.194} & {-0.306} \\
\textbf{Q32Bc $\rightarrow$ 4o} & {\textbf{0.103}} & {0.104} & {-0.951}
                     & {\textbf{0.261}} & {0.268} & {-2.826}
                     & {\textbf{0.223}} & {0.238} & {-6.252} \\
\textbf{Q72B $\rightarrow$ 4o} & {\textbf{0.088}} & {0.092} & {-4.328}
                    & {\textbf{0.242}} & {0.253} & {-4.169}
                    & {\textbf{0.173}} & {0.195} & {-11.454} \\
\midrule
\textbf{Average} & {\textbf{0.127}} & {0.130} & {-1.188}
                 & {\textbf{0.277}} & {0.283} & {-1.897}
                 & {\textbf{0.205}} & {0.212} & {-3.193} \\
\bottomrule
\end{tabular}
\end{adjustbox}
\end{table}

\begin{table}[t]
\centering
\small
\caption{Change in overall cascade loss on the test set when using early abstention (``Early'') 
compared to final-model abstention (``Final''): TriviaQA, TruthfulQA, and XSum.}
\label{tab:overall-loss-2}
\begin{adjustbox}{max width=\textwidth}
\begin{tabular}{lrrrrrrrrrr}
\toprule
\multirow{2}{*}{\textbf{Cascade}} 
 & \multicolumn{3}{c}{\textbf{TriviaQA}}
 & \multicolumn{3}{c}{\textbf{TruthfulQA}}
 & \multicolumn{3}{c}{\textbf{XSum}} \\
 & \textbf{Early} & \textbf{Final} & \textbf{\%$\boldsymbol{\Delta}$}
 & \textbf{Early} & \textbf{Final} & \textbf{\%$\boldsymbol{\Delta}$}
 & \textbf{Early} & \textbf{Final} & \textbf{\%$\boldsymbol{\Delta}$} \\
\midrule
\textbf{L1B $\rightarrow$ L3B}
& {0.275} & {\textbf{0.270}} & {1.594}
& {\textbf{0.428}} & {0.428} & {-0.050}
& {\textbf{0.488}} & {0.497} & {-1.731} \\

\textbf{L1B $\rightarrow$ L8B} 
& {0.170} & {\textbf{0.169}} & {0.549}
& {0.398} & {\textbf{0.395}} & {0.670}
& {\textbf{0.369}} & {0.371} & {-0.519} \\

\textbf{L1B $\rightarrow$ L70B} 
& {0.078} & {\textbf{0.075}} & {4.359}
& {\textbf{0.323}} & {0.327} & {-1.052}
& {\textbf{0.231}} & {0.248} & {-7.131} \\

\textbf{L1B $\rightarrow$ L405B} 
& {0.087} & {\textbf{0.086}} & {1.259}
& {\textbf{0.286}} & {0.298} & {-4.004}
& {\textbf{0.356}} & {0.471} & {-24.317} \\

\textbf{L3B $\rightarrow$ L8B} 
& {0.170} & {\textbf{0.170}} & {0.194}
& {\textbf{0.385}} & {0.387} & {-0.458}
& {\textbf{0.367}} & {0.371} & {-1.152} \\

\textbf{L3B $\rightarrow$ L70B} 
& {\textbf{0.075}} & {0.075} & {-0.153}
& {\textbf{0.323}} & {0.326} & {-0.873}
& {\textbf{0.229}} & {0.246} & {-6.974} \\

\textbf{L3B $\rightarrow$ L405B} 
& {\textbf{0.084}} & {0.086} & {-2.612}
& {\textbf{0.289}} & {0.300} & {-3.782}
& {\textbf{0.354}} & {0.467} & {-24.184} \\

\textbf{L8B $\rightarrow$ L70B} 
& {\textbf{0.080}} & {0.081} & {-1.880}
& {\textbf{0.326}} & {0.329} & {-0.940}
& {\textbf{0.231}} & {0.250} & {-7.552} \\

\textbf{L8B $\rightarrow$ L405B} 
& {\textbf{0.080}} & {0.081} & {-1.014}
& {\textbf{0.290}} & {0.303} & {-4.451}
& {\textbf{0.303}} & {0.381} & {-20.432} \\

\textbf{L70B $\rightarrow$ L405B} 
& {0.069} & {\textbf{0.069}} & {0.005}
& {\textbf{0.296}} & {0.304} & {-2.655}
& {\textbf{0.231}} & {0.249} & {-7.281} \\

\textbf{4o-Mini $\rightarrow$ Q32Bc} 
& {\textbf{0.089}} & {0.091} & {-1.952}
& {\textbf{0.318}} & {0.319} & {-0.363}
& {\textbf{0.054}} & {0.055} & {-1.694} \\

\textbf{4o-Mini $\rightarrow$ Q72B} 
& {0.083} & {\textbf{0.076}} & {9.957}
& {\textbf{0.322}} & {0.322} & {-0.042}
& {0.055} & {\textbf{0.054}} & {0.793} \\

\textbf{4o-Mini $\rightarrow$ 4o} 
& {0.065} & {\textbf{0.056}} & {16.892}
& {\textbf{0.283}} & {0.290} & {-2.537}
& {\textbf{0.054}} & {0.054} & {-1.340} \\

\textbf{Q32Bc $\rightarrow$ Q72B} 
& {0.127} & {\textbf{0.124}} & {2.270}
& {\textbf{0.315}} & {0.316} & {-0.531}
& {\textbf{0.230}} & {0.232} & {-0.850} \\

\textbf{Q32Bc $\rightarrow$ 4o} 
& {0.082} & {\textbf{0.081}} & {1.707}
& {\textbf{0.261}} & {0.270} & {-3.525}
& {\textbf{0.292}} & {0.330} & {-11.391} \\

\textbf{Q72B $\rightarrow$ 4o} 
& {0.061} & {\textbf{0.061}} & {0.776}
& {\textbf{0.280}} & {0.287} & {-2.568}
& {\textbf{0.143}} & {0.145} & {-1.725} \\
\midrule
\textbf{Average}
& {0.105} & {\textbf{0.103}} & {1.997}
& {\textbf{0.320}} & {0.325} & {-1.698}
& {\textbf{0.249}} & {0.276} & {-7.343} \\
\bottomrule
\end{tabular}
\end{adjustbox}
\end{table}

\begin{figure*}[ht!]
    \centering
    \begin{subfigure}[b]{0.32\textwidth}
        \centering
        \includegraphics[width=\textwidth]{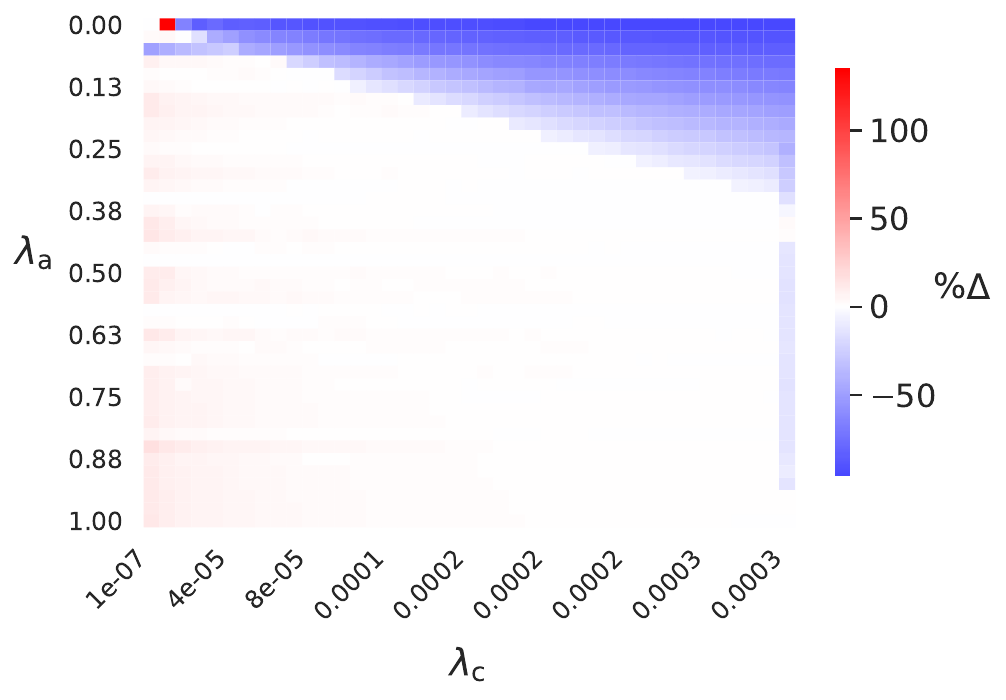}
        \caption{GSM8K}
        \label{fig:sub5}
    \end{subfigure}
    \hfill
    \begin{subfigure}[b]{0.32\textwidth}
        \centering
        \includegraphics[width=\textwidth]{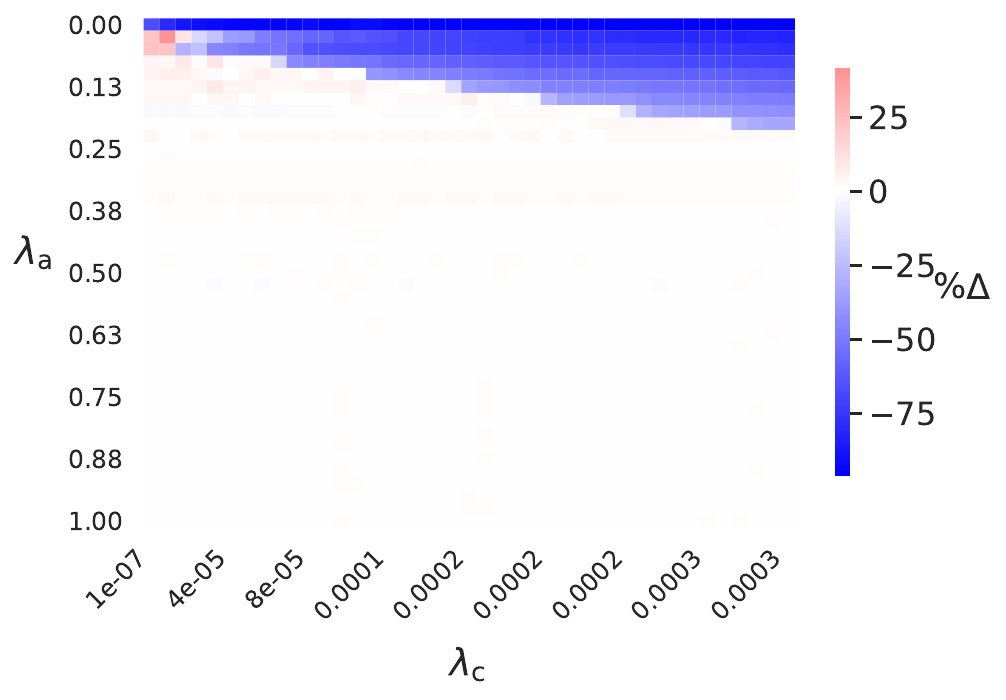}
        \caption{MedMCQA}
        \label{fig:sub2}
    \end{subfigure}
    \hfill
    \begin{subfigure}[b]{0.32\textwidth}
        \centering
        \includegraphics[width=\textwidth]{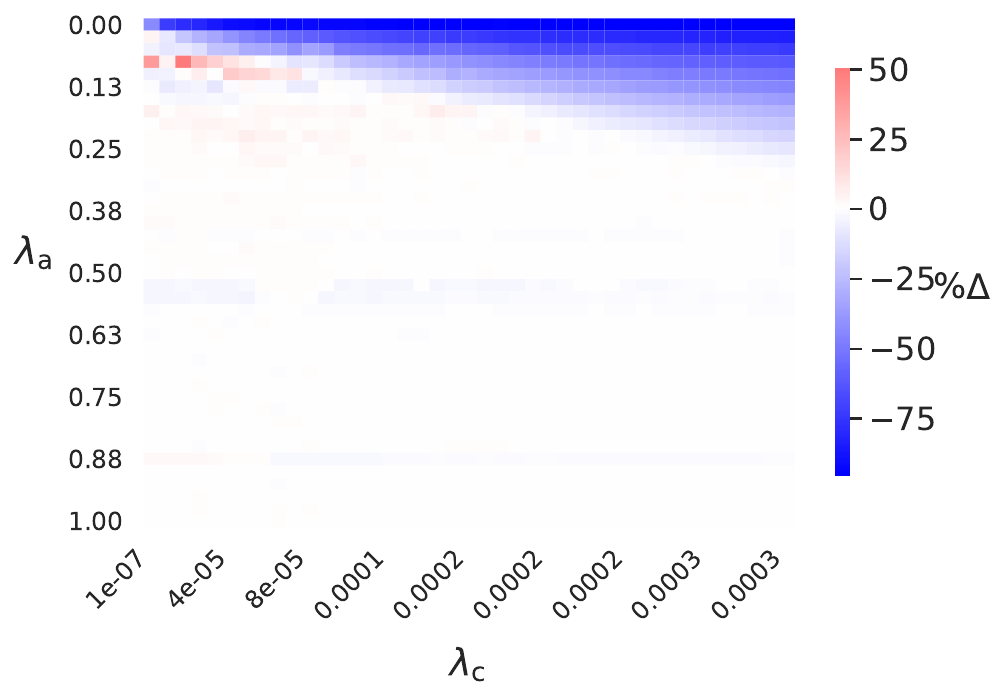}
        \caption{MMLU}
        \label{fig:sub1}
    \end{subfigure}

    \vspace{1em}  

    \begin{subfigure}[b]{0.32\textwidth}
        \centering
        \includegraphics[width=\textwidth]{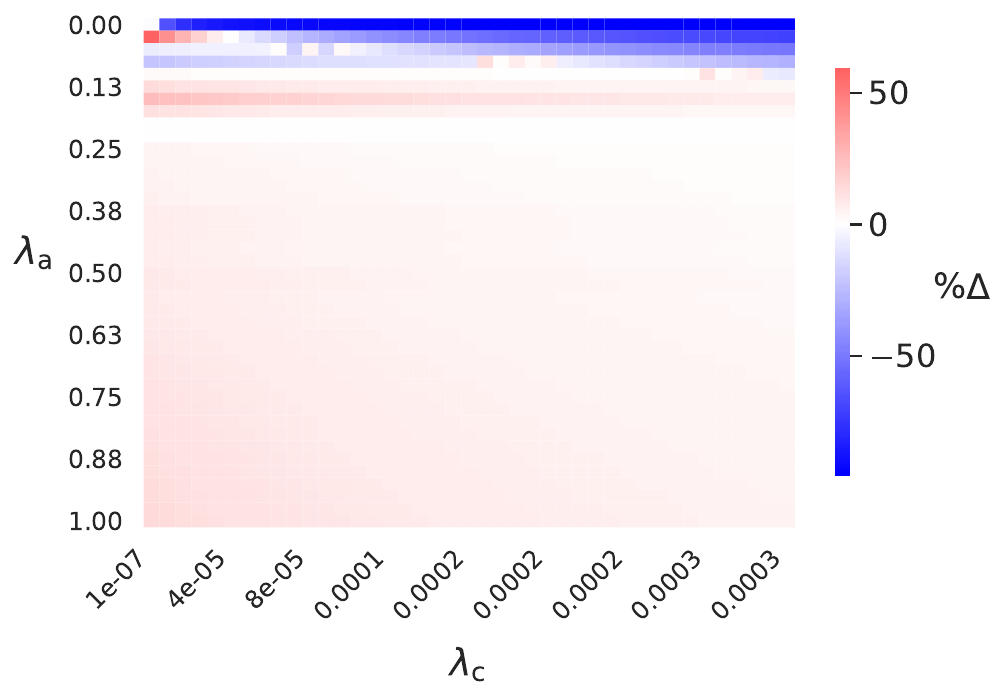}
        \caption{TriviaQA}
        \label{fig:sub3}
    \end{subfigure}
    \hfill
    \begin{subfigure}[b]{0.32\textwidth}
        \centering
        \includegraphics[width=\textwidth]{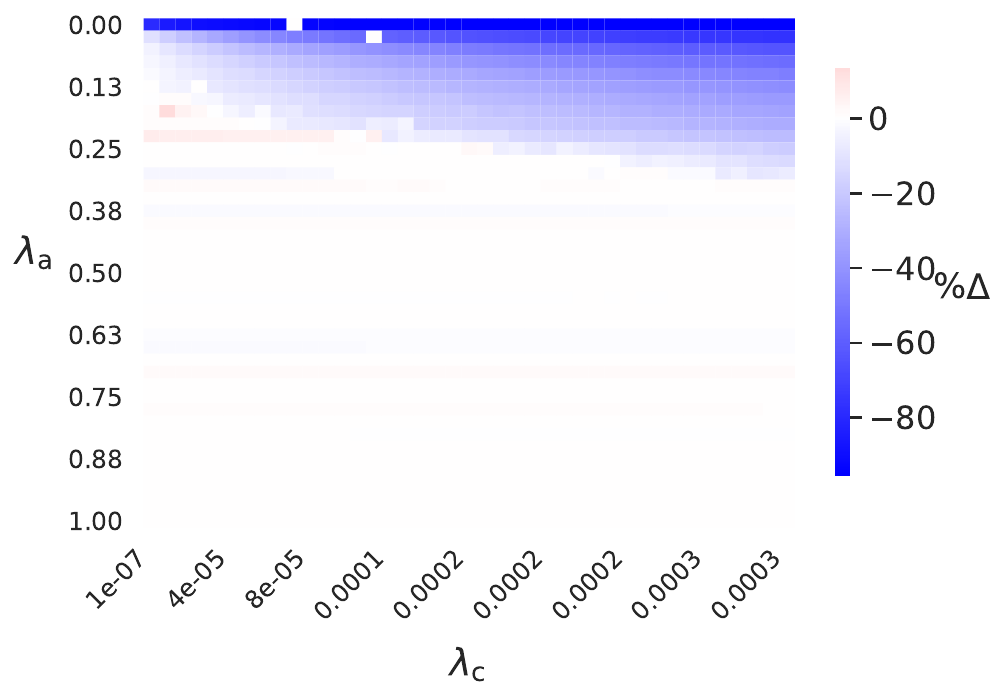}
        \caption{TruthfulQA}
        \label{fig:sub6}
    \end{subfigure}
    \hfill
    \begin{subfigure}[b]{0.32\textwidth}
        \centering
        \includegraphics[width=\textwidth]{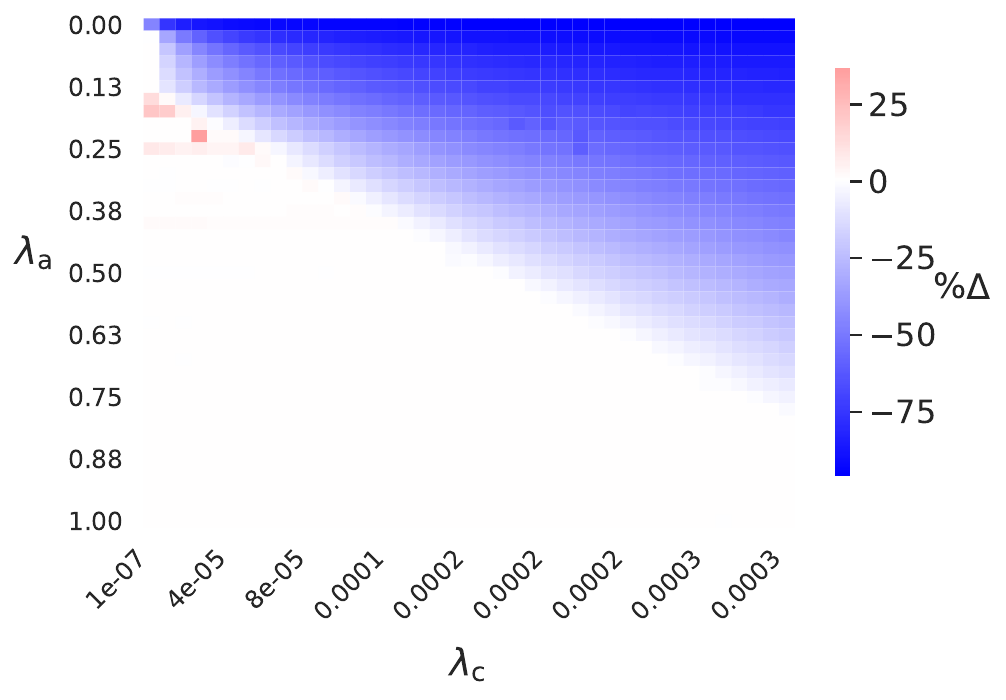}
        \caption{XSum}
        \label{fig:sub4}
    \end{subfigure}

\caption{Percentage change in overall cascade loss when allowing early abstention vs only abstaining at the final model, for the Llama3.2 1B $\rightarrow$ Llama3.1 405B cascade. The performance benefits of early abstention concentrate in the top right corner of the user preference space (blue), where the user's sensitivity to abstention ($\lambda_a$) is low to moderate and the sensitivity to cost ($\lambda_c$) is moderate to high.}
    \label{fig:overall_test_loss}
\end{figure*}

\noindent \textbf{Early abstention reduces overall test loss.} Tables \ref{tab:overall-loss-1} and \ref{tab:overall-loss-2} show the percentage changes in overall cascade loss when allowing the small model in a two-model cascade to abstain. On six benchmarks, allowing early abstention lowers the overall cascade loss (\ref{eq:overall_cascade_loss}) by 2.2\% on average. On five out of the six benchmarks (GSM8K, MedMCQA, MMLU, XSum, and TruthfulQA), early abstention reduces the overall loss by 3.1\% on average, ranging from -1.2\% on GSM8K to -7.3\% on XSum. Only on TriviaQA, early abstention performed worse, raising the overall loss by 2.0\%.

\noindent \textbf{The benefits are concentrated in the upper right quadrant of user preference space.} Figure \ref{fig:overall_test_loss} shows heatmaps of the percentage changes in overall cascade loss across the user preference space $(\lambda_\text{cost}, \lambda_\text{abs})$, for the cascade Llama3.2 1B $\rightarrow$ Llama3.1 405B. The figure illustrates that the benefits of early abstention are highly concentrated in the upper right quadrant of the user preference space, where cost is a concern ($\lambda_c$ moderate to high) and the penalty on abstention is not severe ($\lambda_a$ low to moderate). This trend holds generally across cascades.

\noindent \textbf{Early abstention makes more effective use of the ability to abstain.} Tables \ref{tab:early-abstention-by-benchmark} and \ref{tab:early-abstention-by-cascade} show the effects of early abstention on the test error rate, expected cost, and abstention rate, broken down by benchmark and by cascade. These results show that allowing the small model in a two-model cascade to directly abstain on a query leads to significant cost reductions (-13\% on average), and error reductions (-5\% on average), at the expense of a moderate increase in the abstention rate (+4\% on average).\footnote{We report the change in abstention rate as a difference rather than a percentage change because early abstention sometimes uses low amounts of abstention when final-model abstention uses none, leading to undefined percentage changes.} Our interpretation is that early abstention makes more effective use of the ability to abstain, leading a cascade to make greater use of abstention in order to improve overall performance. Appendix D provides more detailed results on the changes in the test error rate, expected cost, and abstention rate due to early abstention.

\begin{table}[ht]
    \centering
    \caption{Early abstention trades a 4\% absolute increase in the abstention rate for a 5\% error reduction and 13\% cost reduction, on average. Results broken down by benchmark.}
    \label{tab:early-abstention-by-benchmark}
        \begin{tabular}{lrrr}
            \toprule
            \textbf{Benchmark} 
            & {$\boldsymbol{\%\Delta}$\textbf{Error}}
            & {$\boldsymbol{\%\Delta}$\textbf{Cost}} 
            & {$\boldsymbol{\Delta }$\textbf{Abstention}} \\
            \midrule
            GSM8K       & -6.763  & -4.097   & 0.037 \\
            MedMCQA     & -1.592  & -11.085  & 0.022 \\
            MMLU        & -3.175  & -15.178  & 0.030 \\
            TriviaQA    & -1.360  & -7.512   & 0.020 \\
            TruthfulQA  & -2.380  & -19.713  & 0.025 \\
            XSum        & -14.635 & -20.273  & 0.110 \\
            \midrule
            \textbf{Average} 
                        & -4.984  & -12.976  & 0.041 \\
            \bottomrule
        \end{tabular}
\end{table}

\begin{table}[ht]
    \centering
    \caption{Early abstention trades a 4\% absolute increase in the abstention rate for a 5\% error reduction and 13\% cost reduction, on average. Results broken down by cascade.}
    \label{tab:early-abstention-by-cascade}
        \begin{tabular}{lrrr}
            \toprule
            \textbf{Cascade} 
            & \(\boldsymbol{\%\Delta \textbf{Error}}\)
            & \(\boldsymbol{\%\Delta \textbf{Cost}}\) 
            & \(\boldsymbol{\Delta \textbf{Abstention}}\) \\
            \midrule
            L1B \(\to\) L3B          & -3.581   & -13.764  & 0.022 \\
            L1B \(\to\) L8B          & -0.618   & -12.273  & 0.013 \\
            L1B \(\to\) L70B         & -2.483   & -12.292  & 0.042 \\
            L1B \(\to\) L405B        & -10.650  & -21.361  & 0.112 \\
            L3B \(\to\) L8B          & -2.205   & -11.952  & 0.016 \\
            L3B \(\to\) L70B         & -1.597   & -12.765  & 0.023 \\
            L3B \(\to\) L405B        & -10.352  & -21.551  & 0.101 \\
            L8B \(\to\) L70B         & -2.272   & -11.803  & 0.020 \\
            L8B \(\to\) L405B        & -9.660   & -17.936  & 0.081 \\
            L70B \(\to\) L405B       & -2.736   & -12.524  & 0.022 \\
            4o-Mini \(\to\) Q32Bc     & -5.710   & -12.002  & 0.017 \\
            4o-Mini \(\to\) Q72B     & -1.127   & -7.993   & 0.005 \\
            4o-Mini \(\to\) 4o       & -3.908   & -14.766  & 0.031 \\
            Q32Bc \(\to\) Q72B        & -4.574   & -0.235   & 0.022 \\
            Q32Bc \(\to\) 4o          & -9.424   & -14.039  & 0.077 \\
            Q72B \(\to\) 4o          & -8.848   & -10.368  & 0.046 \\
            \midrule
            \textbf{Average}         & -4.984   & -12.976  & 0.041 \\
            \bottomrule
        \end{tabular}
\end{table}

\section{Conclusion}

In this work, we investigated the question of early abstention in cascade design: when designing an LLM cascade capable of abstaining from difficult queries, does it make sense to allow small models earlier in the cascade to abstain, compared to only abstaining at the final model? Our results indicate that early abstention improves overall performance, as measured by a 2.2\% average reduction in overall cascade loss across six language modeling benchmarks. We find that the benefits are concentrated in the upper right quadrant of the user preference space, where cost is a concern and abstention is not heavily penalized. Overall, early abstention allows a cascade to make more effective use of the ability to abstain, trading a moderate increase in the abstention rate (+4.1\% on average) for significant cost savings (-13.0\% on average) and some error reduction (-5.0\% on average).

These positive results can be explained by the correlations between the error patterns of small and large LLMs. On one level, our findings show that these correlations can be leveraged to improve the performance of an LLM system with abstention. Viewed more broadly, an LLM cascade with early abstention is an example of an \textit{integrated} LLM system that incorporates interactions between models to improve performance. As the LLM research community increasingly moves towards ensembles of LLMs collaborating with each other to solve tasks more efficiently and reliably, we expect that leveraging interactions between models will become critical for optimizing LLM systems.

\bibliography{main}

\newpage

\section*{Appendix A: Modified Equations for Performance Metrics}

The ``rational tuning'' approach towards optimizing cascade thresholds (\citealp{zellinger2025}) relies on differentiable expressions for the probability of correctness and the expected cost. We extend this approach to include abstention, in accordance with our cascade model (\ref{eq:cascade_model}). The modified equations for $\mathbb{P}(\text{Correct})$, $\mathbb{E}[\text{Cost}]$, and $\mathbb{P}(\text{Abstention})$ become
\begin{align*}
    \mathbb{P}(\text{Correct}) = & ~Q_1 + \sum_{i=2}^{k} P_1 \left( \prod_{j=2}^{i-1} P_{j-1, j} \right) Q_{i-1,i} \\
    \mathbb{E}[\text{Cost}] = & ~(1-P_1) ~\mathbb{E}[C_1] + \sum_{i=2}^{k} P_1 \left( \prod_{j=2}^{i-1} P_{j-1,j} \right) (1 - P_{i-1,i}) \left(\sum_{j=1}^{i} \mathbb{E}[C_j] \right) \\
    \mathbb{P}(\text{Abstention}) = & ~\mathbb{P}(\Phi_1 < \xi_1) + \sum_{i=2}^{k} P_1 \left( \prod_{j=2}^{i-1} P_{j-1,j} \right) \mathbb{P}(\Phi_i < \xi_i | \Phi_{i-1} \in [\xi_{i-1},\phi_{i-1}]),
\end{align*}
where 
\begin{align*}
    Q_1 := & \int_{\{\Phi_1 > \phi_1\}} \Phi_1  \text{~d}\mathbb{P} \\
    Q_{i,j} := & \int_{\{\Phi_j > \phi_j\}} \Phi_j(\omega)  \text{~d}\mathbb{P}(\omega | \Phi_{i} \in [\xi_i, \phi_{i}]) \\
    P_1 := & ~\mathbb{P}(\Phi_1 \in [\xi_1, \phi_1]) \\
    P_{i,j} := & ~\mathbb{P}(\Phi_j \in [\xi_j, \phi_j] ~|~ \Phi_i \in [\xi_i, \phi_i]), \\
\end{align*}
and $\mathbb{E}[C_i]$ is the expected cost per query of $M_i$. As in Proposition 2 of \cite{zellinger2025}, we let $\phi_k := -\infty$, even though the final model does not have a deferral threshold.

\section*{Appendix B: Logistic Regression for Abstention Prediction}

To predict the final model's decision to abstain, we fit a logistic regression model based on the earlier model's confidence scores.

Consider a cascade $C = M_1 \rightarrow ... \rightarrow M_k$ and let the observed confidence scores of the final model $M_k$ on the training set be $\{ \Phi_k^{(i)} \}_{i=1}^{n}$. To achieve abstention rates of 20\% and 30\%, we set the abstention threshold $\xi_k$ to be the lower 20\% and 30\% quantiles of $\{ \Phi_k^{(i)} \}_{i=1}^{n}$. Applying this abstention policy leads to abstention decisions $y_i = \mathds{1}[\Phi_k^{(i)} < \xi_k]$ for $i=1, ..., n$. Using the $y_i$ as ground truth labels, we fit the following logistic regression model for the probability $p$ that the final model abstains:
\begin{equation}
    \log \left( \frac{p}{1-p} \right) = \beta_0 + \sum_{j=1}^{k-1} \beta_j f(\Phi_j^{(i)}) + \epsilon,
\end{equation}
where $f$ is a nonlinear transformation of the confidence scores $\Phi_j^{(i)}$. Specifically, $f(\Phi_j^{(i)})$ is the ``transformed'' raw confidence signal as in \cite{zellinger2025}. For example, on multiple-choice QA benchmarks, 
\begin{equation}
    f(\Phi) = \log \left( \frac{1}{1-p_\text{raw}} \right),
\end{equation}
where $p_\text{raw}$ is maximum conditional next-token probability among the valid answer choices (e.g., ``A'', ``B'', ``C'', or ``D'' on MMLU).

\section*{Appendix C: Smoothing Procedure for Optimal Cascade Thresholds}

We apply light smoothing on the grid of optimal cascade thresholds to avoid potential artifacts from numerical optimization. Our procedure consists of two steps:
\begin{itemize}
    \item Step 1: Identify outliers
    \item Step 2: Replace outliers by the mean values of their neighbors
\end{itemize}
In Step 1, we call a threshold vector $\boldsymbol{\theta}$ an \textit{outlier} if $\overline{\theta} := \sum_{i=1}^{n} \theta_i$ satisfies
\begin{equation}
    (\overline{\theta} - \frac{1}{|N|}\sum_{n \in N} \overline{\theta_n})^2 > r \text{Var}(\overline{\theta_n}, n \in N),
\end{equation}
where $n \in N$ indexes the set of neighboring optimal thresholds on the 2D grid (an interior point has four neighbors).

Reported results in the paper use $r=10$. This results in smoothing 3.5\% of data points for early abstention and 2.7\% of data points for final-model abstention.

\section*{Appendix D: Detailed Changes in Error, Cost, and Abstention}

Below we provide detailed results on the changes in the test error rate (Table \ref{tab:error-test}), expected cost (Table \ref{tab:cost-test}), and abstention rate (Table \ref{tab:abstention-test}) resulting from using early abstention, compared to abstaining only at the final model.

\begin{table}[h]
\centering
\small
\caption{Change in \textbf{error rate} on the test set when using early abstention (``Early'') compared to final-model abstention (``Final''). Bolded numbers indicate outperformance.}
\label{tab:error-test}
\begin{adjustbox}{max width=\textwidth}
\begin{tabular}{lrrrrrrrrrrrrrrrrrrrrr}
\toprule
\multirow{2}{*}{\textbf{Cascade}} & \multicolumn{3}{c}{\textbf{GSM8K}} & \multicolumn{3}{c}{\textbf{MedMCQA}} & \multicolumn{3}{c}{\textbf{MMLU}} & \multicolumn{3}{c}{\textbf{TriviaQA}} & \multicolumn{3}{c}{\textbf{TruthfulQA}} & \multicolumn{3}{c}{\textbf{XSum}} & \multicolumn{3}{c}{\textbf{Average}} \\
 & \textbf{Early} & \textbf{Final} & \textbf{$\boldsymbol{\%\Delta}$} & \textbf{Early} & \textbf{Final} & \textbf{$\boldsymbol{\%\Delta}$} & \textbf{Early} & \textbf{Final} & \textbf{$\boldsymbol{\%\Delta}$} & \textbf{Early} & \textbf{Final} & \textbf{$\boldsymbol{\%\Delta}$} & \textbf{Early} & \textbf{Final} & \textbf{$\boldsymbol{\%\Delta}$} & \textbf{Early} & \textbf{Final} & \textbf{$\boldsymbol{\%\Delta}$} & \textbf{Early} & \textbf{Final} & \textbf{$\boldsymbol{\%\Delta}$} \\
\midrule
\textbf{L1B $\rightarrow$ L3B} & {\textbf{0.168}} & {0.174} & {-3.108} & {0.228} & {\textbf{0.225}} & {1.620} & {\textbf{0.203}} & {0.214} & {-5.281} & {\textbf{0.177}} & {0.182} & {-2.755} & {0.248} & {\textbf{0.246}} & {0.810} & {\textbf{0.148}} & {0.170} & {-12.773} & {\textbf{0.195}} & {0.202} & {-3.581} \\
\textbf{L1B $\rightarrow$ L8B} & {\textbf{0.130}} & {0.131} & {-0.368} & {\textbf{0.239}} & {0.240} & {-0.266} & {\textbf{0.171}} & {0.173} & {-1.265} & {\textbf{0.130}} & {0.136} & {-4.117} & {0.279} & {\textbf{0.275}} & {1.264} & {0.189} & {\textbf{0.187}} & {1.044} & {\textbf{0.190}} & {0.190} & {-0.618} \\
\textbf{L1B $\rightarrow$ L70B} & {\textbf{0.045}} & {0.047} & {-3.877} & {0.187} & {\textbf{0.186}} & {0.355} & {\textbf{0.111}} & {0.111} & {-0.048} & {0.055} & {\textbf{0.055}} & {0.267} & {\textbf{0.221}} & {0.226} & {-2.175} & {\textbf{0.110}} & {0.122} & {-9.421} & {\textbf{0.122}} & {0.125} & {-2.483} \\
\textbf{L1B $\rightarrow$ L405B} & {\textbf{0.022}} & {0.027} & {-19.898} & {\textbf{0.177}} & {0.184} & {-3.992} & {0.093} & {\textbf{0.092}} & {0.425} & {\textbf{0.039}} & {0.040} & {-0.878} & {\textbf{0.185}} & {0.193} & {-4.365} & {\textbf{0.085}} & {0.131} & {-35.191} & {\textbf{0.100}} & {0.111} & {-10.650} \\
\textbf{L3B $\rightarrow$ L8B} & {\textbf{0.159}} & {0.169} & {-5.530} & {\textbf{0.240}} & {0.246} & {-2.399} & {\textbf{0.174}} & {0.176} & {-0.987} & {\textbf{0.137}} & {0.138} & {-0.543} & {\textbf{0.263}} & {0.270} & {-2.539} & {\textbf{0.188}} & {0.191} & {-1.235} & {\textbf{0.194}} & {0.198} & {-2.205} \\
\textbf{L3B $\rightarrow$ L70B} & {\textbf{0.077}} & {0.079} & {-2.704} & {0.192} & {\textbf{0.190}} & {1.061} & {0.112} & {\textbf{0.111}} & {1.201} & {0.056} & {\textbf{0.055}} & {0.561} & {\textbf{0.223}} & {0.227} & {-1.617} & {\textbf{0.109}} & {0.119} & {-8.083} & {\textbf{0.128}} & {0.130} & {-1.597} \\
\textbf{L3B $\rightarrow$ L405B} & {\textbf{0.132}} & {0.150} & {-11.961} & {\textbf{0.190}} & {0.204} & {-6.860} & {\textbf{0.094}} & {0.096} & {-2.441} & {0.040} & {\textbf{0.040}} & {0.334} & {\textbf{0.185}} & {0.196} & {-5.355} & {\textbf{0.093}} & {0.145} & {-35.832} & {\textbf{0.122}} & {0.138} & {-10.352} \\
\textbf{L8B $\rightarrow$ L70B} & {\textbf{0.068}} & {0.068} & {-0.726} & {0.203} & {\textbf{0.197}} & {3.173} & {\textbf{0.115}} & {0.116} & {-1.131} & {\textbf{0.061}} & {0.063} & {-3.258} & {\textbf{0.222}} & {0.226} & {-1.468} & {\textbf{0.116}} & {0.129} & {-10.222} & {\textbf{0.131}} & {0.133} & {-2.272} \\
\textbf{L8B $\rightarrow$ L405B} & {\textbf{0.099}} & {0.105} & {-5.294} & {\textbf{0.206}} & {0.217} & {-5.216} & {\textbf{0.095}} & {0.099} & {-3.848} & {\textbf{0.045}} & {0.046} & {-1.356} & {\textbf{0.183}} & {0.200} & {-8.456} & {\textbf{0.138}} & {0.209} & {-33.793} & {\textbf{0.128}} & {0.146} & {-9.660} \\
\textbf{L70B $\rightarrow$ L405B} & {0.046} & {\textbf{0.046}} & {0.305} & {\textbf{0.193}} & {0.197} & {-2.343} & {\textbf{0.099}} & {0.099} & {-0.259} & {0.049} & {\textbf{0.047}} & {3.788} & {\textbf{0.197}} & {0.200} & {-1.552} & {\textbf{0.112}} & {0.134} & {-16.357} & {\textbf{0.116}} & {0.121} & {-2.736} \\
\textbf{4o-Mini $\rightarrow$ Q32Bc} & {\textbf{0.047}} & {0.051} & {-8.602} & {\textbf{0.201}} & {0.203} & {-0.721} & {0.160} & {\textbf{0.159}} & {0.406} & {\textbf{0.062}} & {0.069} & {-10.337} & {\textbf{0.237}} & {0.244} & {-2.602} & {\textbf{0.022}} & {0.025} & {-12.404} & {\textbf{0.122}} & {0.125} & {-5.710} \\
\textbf{4o-Mini $\rightarrow$ Q72B} & {\textbf{0.050}} & {0.053} & {-4.445} & {0.200} & {\textbf{0.199}} & {0.455} & {0.138} & {\textbf{0.135}} & {1.878} & {0.057} & {\textbf{0.056}} & {0.963} & {\textbf{0.241}} & {0.242} & {-0.671} & {\textbf{0.024}} & {0.025} & {-4.941} & {\textbf{0.118}} & {0.118} & {-1.127} \\
\textbf{4o-Mini $\rightarrow$ 4o} & {\textbf{0.058}} & {0.067} & {-13.938} & {\textbf{0.153}} & {0.155} & {-1.153} & {\textbf{0.107}} & {0.108} & {-1.229} & {0.038} & {\textbf{0.038}} & {2.280} & {\textbf{0.204}} & {0.209} & {-2.449} & {\textbf{0.020}} & {0.022} & {-6.962} & {\textbf{0.097}} & {0.100} & {-3.908} \\
\textbf{Q32Bc $\rightarrow$ Q72B} & {\textbf{0.042}} & {0.049} & {-14.945} & {\textbf{0.195}} & {0.196} & {-0.644} & {0.137} & {\textbf{0.137}} & {0.109} & {\textbf{0.076}} & {0.077} & {-1.301} & {\textbf{0.219}} & {0.222} & {-1.486} & {\textbf{0.061}} & {0.067} & {-9.177} & {\textbf{0.122}} & {0.125} & {-4.574} \\
\textbf{Q32Bc $\rightarrow$ 4o} & {\textbf{0.048}} & {0.049} & {-1.889} & {\textbf{0.142}} & {0.148} & {-3.676} & {\textbf{0.158}} & {0.186} & {-15.105} & {\textbf{0.022}} & {0.023} & {-2.698} & {\textbf{0.163}} & {0.167} & {-2.615} & {\textbf{0.115}} & {0.165} & {-30.562} & {\textbf{0.108}} & {0.123} & {-9.424} \\
\textbf{Q72B $\rightarrow$ 4o} & {\textbf{0.037}} & {0.041} & {-11.229} & {\textbf{0.162}} & {0.171} & {-4.863} & {\textbf{0.126}} & {0.165} & {-23.230} & {\textbf{0.031}} & {0.032} & {-2.710} & {\textbf{0.175}} & {0.180} & {-2.803} & {\textbf{0.039}} & {0.043} & {-8.251} & {\textbf{0.095}} & {0.105} & {-8.848} \\
\midrule
\textbf{Average} & {\textbf{0.077}} & {0.082} & {-6.763} & {\textbf{0.194}} & {0.197} & {-1.592} & {\textbf{0.131}} & {0.136} & {-3.175} & {\textbf{0.067}} & {0.069} & {-1.360} & {\textbf{0.215}} & {0.220} & {-2.380} & {\textbf{0.098}} & {0.118} & {-14.635} & {\textbf{0.130}} & {0.137} & {-4.984} \\
\bottomrule
\end{tabular}
\end{adjustbox}
\end{table}

\begin{table}[h]
\centering
\small
\caption{Change in \textbf{expected cost} on the test set when using early abstention (``Early'') compared to final-model abstention (``Final''). Bolded numbers indicate outperformance.}
\label{tab:cost-test}
\begin{adjustbox}{max width=\textwidth}
\begin{tabular}{lrrrrrrrrrrrrrrrrrrrrr}
\toprule
\multirow{2}{*}{\textbf{Cascade}} & \multicolumn{3}{c}{\textbf{GSM8K}} & \multicolumn{3}{c}{\textbf{MedMCQA}} & \multicolumn{3}{c}{\textbf{MMLU}} & \multicolumn{3}{c}{\textbf{TriviaQA}} & \multicolumn{3}{c}{\textbf{TruthfulQA}} & \multicolumn{3}{c}{\textbf{XSum}} & \multicolumn{3}{c}{\textbf{Average}} \\
 & \textbf{Early} & \textbf{Final} & \textbf{$\boldsymbol{\%\Delta}$} & \textbf{Early} & \textbf{Final} & \textbf{$\boldsymbol{\%\Delta}$} & \textbf{Early} & \textbf{Final} & \textbf{$\boldsymbol{\%\Delta}$} & \textbf{Early} & \textbf{Final} & \textbf{$\boldsymbol{\%\Delta}$} & \textbf{Early} & \textbf{Final} & \textbf{$\boldsymbol{\%\Delta}$} & \textbf{Early} & \textbf{Final} & \textbf{$\boldsymbol{\%\Delta}$} & \textbf{Early} & \textbf{Final} & \textbf{$\boldsymbol{\%\Delta}$} \\
\midrule
\textbf{L1B $\rightarrow$ L3B} & {\textbf{71.601}} & {73.988} & {-3.226} & {\textbf{31.986}} & {33.354} & {-4.102} & {\textbf{34.008}} & {37.118} & {-8.379} & {\textbf{14.860}} & {16.966} & {-12.417} & {\textbf{15.373}} & {19.043} & {-19.269} & {\textbf{86.081}} & {132.816} & {-35.188} & {\textbf{42.318}} & {52.214} & {-13.764} \\
\textbf{L1B $\rightarrow$ L8B} & {\textbf{104.607}} & {111.371} & {-6.073} & {\textbf{47.126}} & {49.997} & {-5.741} & {\textbf{50.688}} & {55.466} & {-8.614} & {\textbf{23.236}} & {25.642} & {-9.382} & {\textbf{23.479}} & {30.187} & {-22.221} & {\textbf{159.931}} & {204.010} & {-21.606} & {\textbf{68.178}} & {79.445} & {-12.273} \\
\textbf{L1B $\rightarrow$ L70B} & {\textbf{354.301}} & {380.237} & {-6.821} & {\textbf{155.516}} & {165.923} & {-6.272} & {\textbf{162.822}} & {185.591} & {-12.268} & {\textbf{80.017}} & {84.741} & {-5.574} & {\textbf{79.822}} & {101.761} & {-21.559} & {\textbf{550.516}} & {699.117} & {-21.256} & {\textbf{230.499}} & {269.562} & {-12.292} \\
\textbf{L1B $\rightarrow$ L405B} & {\textbf{980.892}} & {1192.188} & {-17.723} & {\textbf{441.203}} & {507.625} & {-13.085} & {\textbf{456.881}} & {561.538} & {-18.638} & {\textbf{253.605}} & {275.737} & {-8.026} & {\textbf{245.452}} & {322.141} & {-23.806} & {\textbf{1144.376}} & {2154.710} & {-46.890} & {\textbf{587.068}} & {835.657} & {-21.361} \\
\textbf{L3B $\rightarrow$ L8B} & {\textbf{52.641}} & {56.314} & {-6.523} & {\textbf{39.131}} & {42.783} & {-8.536} & {\textbf{44.607}} & {50.427} & {-11.541} & {\textbf{24.214}} & {25.324} & {-4.382} & {\textbf{22.071}} & {28.303} & {-22.017} & {\textbf{164.680}} & {202.595} & {-18.715} & {\textbf{57.891}} & {67.624} & {-11.952} \\
\textbf{L3B $\rightarrow$ L70B} & {\textbf{245.738}} & {257.410} & {-4.534} & {\textbf{132.535}} & {145.584} & {-8.963} & {\textbf{145.695}} & {171.722} & {-15.156} & {\textbf{81.480}} & {84.531} & {-3.609} & {\textbf{77.356}} & {100.440} & {-22.983} & {\textbf{546.826}} & {695.215} & {-21.344} & {\textbf{204.938}} & {242.483} & {-12.765} \\
\textbf{L3B $\rightarrow$ L405B} & {\textbf{266.372}} & {319.065} & {-16.515} & {\textbf{354.817}} & {413.319} & {-14.154} & {\textbf{386.859}} & {495.710} & {-21.959} & {\textbf{260.629}} & {274.757} & {-5.142} & {\textbf{239.152}} & {319.538} & {-25.157} & {\textbf{1112.024}} & {2073.907} & {-46.380} & {\textbf{436.642}} & {649.383} & {-21.551} \\
\textbf{L8B $\rightarrow$ L70B} & {\textbf{297.188}} & {299.470} & {-0.762} & {\textbf{140.904}} & {154.070} & {-8.546} & {\textbf{144.409}} & {171.579} & {-15.835} & {\textbf{89.032}} & {92.948} & {-4.213} & {\textbf{90.105}} & {113.664} & {-20.727} & {\textbf{541.207}} & {682.801} & {-20.737} & {\textbf{217.141}} & {252.422} & {-11.803} \\
\textbf{L8B $\rightarrow$ L405B} & {\textbf{378.868}} & {386.338} & {-1.934} & {\textbf{344.400}} & {393.865} & {-12.559} & {\textbf{345.498}} & {458.153} & {-24.589} & {\textbf{232.487}} & {238.229} & {-2.410} & {\textbf{250.111}} & {334.041} & {-25.126} & {\textbf{784.729}} & {1330.056} & {-41.000} & {\textbf{389.349}} & {523.447} & {-17.936} \\
\textbf{L70B $\rightarrow$ L405B} & {\textbf{360.270}} & {375.946} & {-4.170} & {\textbf{341.569}} & {368.177} & {-7.227} & {\textbf{280.033}} & {394.310} & {-28.982} & {\textbf{124.742}} & {134.139} & {-7.005} & {\textbf{280.365}} & {337.100} & {-16.830} & {\textbf{758.173}} & {851.186} & {-10.927} & {\textbf{357.525}} & {410.143} & {-12.524} \\
\textbf{4o-Mini $\rightarrow$ Q32Bc} & {\textbf{305.301}} & {317.501} & {-3.843} & {\textbf{80.971}} & {97.703} & {-17.125} & {\textbf{105.827}} & {113.517} & {-6.774} & {\textbf{17.863}} & {21.886} & {-18.381} & {\textbf{54.924}} & {67.327} & {-18.422} & {\textbf{173.952}} & {187.989} & {-7.467} & {\textbf{123.140}} & {134.320} & {-12.002} \\
\textbf{4o-Mini $\rightarrow$ Q72B} & {237.061} & {\textbf{230.590}} & {2.806} & {\textbf{104.568}} & {112.711} & {-7.225} & {\textbf{106.761}} & {113.088} & {-5.594} & {\textbf{22.287}} & {26.170} & {-14.836} & {\textbf{58.558}} & {69.758} & {-16.055} & {\textbf{182.660}} & {196.515} & {-7.051} & {\textbf{118.649}} & {124.805} & {-7.993} \\
\textbf{4o-Mini $\rightarrow$ 4o} & {\textbf{292.712}} & {319.961} & {-8.516} & {\textbf{221.225}} & {254.593} & {-13.106} & {\textbf{181.611}} & {233.275} & {-22.147} & {\textbf{45.837}} & {53.248} & {-13.919} & {\textbf{195.566}} & {249.827} & {-21.719} & {\textbf{269.220}} & {296.455} & {-9.187} & {\textbf{201.029}} & {234.560} & {-14.766} \\
\textbf{Q32Bc $\rightarrow$ Q72B} & {429.898} & {\textbf{363.289}} & {18.335} & {\textbf{256.089}} & {275.515} & {-7.051} & {\textbf{254.571}} & {261.253} & {-2.557} & {\textbf{149.262}} & {151.646} & {-1.572} & {\textbf{136.128}} & {143.954} & {-5.437} & {\textbf{1119.445}} & {1155.584} & {-3.127} & {\textbf{390.899}} & {391.873} & {-0.235} \\
\textbf{Q32Bc $\rightarrow$ 4o} & {\textbf{367.540}} & {372.570} & {-1.350} & {\textbf{468.630}} & {552.128} & {-15.123} & {\textbf{180.103}} & {264.787} & {-31.982} & {\textbf{283.386}} & {297.182} & {-4.642} & {\textbf{310.432}} & {381.989} & {-18.733} & {\textbf{1423.186}} & {1624.714} & {-12.404} & {\textbf{505.546}} & {582.228} & {-14.039} \\
\textbf{Q72B $\rightarrow$ 4o} & {\textbf{349.557}} & {366.776} & {-4.695} & {\textbf{255.560}} & {357.690} & {-28.553} & {\textbf{174.184}} & {188.993} & {-7.836} & {\textbf{134.595}} & {141.211} & {-4.686} & {\textbf{328.934}} & {388.573} & {-15.348} & {\textbf{751.315}} & {759.613} & {-1.092} & {\textbf{332.358}} & {367.143} & {-10.368} \\
\midrule
\textbf{Average} & {\textbf{318.409}} & {338.938} & {-4.097} & {\textbf{213.514}} & {245.315} & {-11.085} & {\textbf{190.910}} & {234.783} & {-15.178} & {\textbf{114.846}} & {121.522} & {-7.512} & {\textbf{150.489}} & {187.978} & {-19.713} & {\textbf{610.520}} & {827.955} & {-20.273} & {\textbf{266.448}} & {326.082} & {-12.976} \\
\bottomrule
\end{tabular}
\end{adjustbox}
\end{table}

\begin{table}[h]
\centering
\small
\caption{Change in \textbf{abstention rate} on the test set when using early abstention (``Early'') compared to final-model abstention (``Final''). Bolded numbers indicate outperformance.}
\label{tab:abstention-test}
\begin{adjustbox}{max width=\textwidth}
\begin{tabular}{lrrrrrrrrrrrrrrrrrrrrr}
\toprule
\multirow{2}{*}{\textbf{Cascade}} & \multicolumn{3}{c}{\textbf{GSM8K}} & \multicolumn{3}{c}{\textbf{MedMCQA}} & \multicolumn{3}{c}{\textbf{MMLU}} & \multicolumn{3}{c}{\textbf{TriviaQA}} & \multicolumn{3}{c}{\textbf{TruthfulQA}} & \multicolumn{3}{c}{\textbf{XSum}} & \multicolumn{3}{c}{\textbf{Average}} \\
 & \textbf{Early} & \textbf{Final} & \textbf{$\boldsymbol{\Delta}$} & \textbf{Early} & \textbf{Final} & \textbf{$\boldsymbol{\Delta}$} & \textbf{Early} & \textbf{Final} & \textbf{$\boldsymbol{\Delta}$} & \textbf{Early} & \textbf{Final} & \textbf{$\boldsymbol{\Delta}$} & \textbf{Early} & \textbf{Final} & \textbf{$\boldsymbol{\Delta}$} & \textbf{Early} & \textbf{Final} & \textbf{$\boldsymbol{\Delta}$} & \textbf{Early} & \textbf{Final} & \textbf{$\boldsymbol{\Delta}$} \\
\midrule
\textbf{L1B $\rightarrow$ L3B} & {0.129} & {\textbf{0.094}} & {0.035} & {\textbf{0.428}} & {0.428} & {-0.000} & {0.434} & {\textbf{0.409}} & {0.025} & {0.407} & {\textbf{0.362}} & {0.045} & {\textbf{0.593}} & {0.594} & {-0.001} & {0.806} & {\textbf{0.777}} & {0.029} & {0.466} & {\textbf{0.444}} & {0.022} \\
\textbf{L1B $\rightarrow$ L8B} & {\textbf{0.126}} & {0.127} & {-0.001} & {0.399} & {\textbf{0.396}} & {0.004} & {0.394} & {\textbf{0.384}} & {0.010} & {0.229} & {\textbf{0.186}} & {0.043} & {0.481} & {\textbf{0.480}} & {0.001} & {0.525} & {\textbf{0.501}} & {0.023} & {0.359} & {\textbf{0.346}} & {0.013} \\
\textbf{L1B $\rightarrow$ L70B} & {0.089} & {\textbf{0.040}} & {0.048} & {0.201} & {\textbf{0.188}} & {0.013} & {0.239} & {\textbf{0.222}} & {0.017} & {0.092} & {\textbf{0.062}} & {0.030} & {0.408} & {\textbf{0.390}} & {0.018} & {0.264} & {\textbf{0.138}} & {0.126} & {0.215} & {\textbf{0.173}} & {0.042} \\
\textbf{L1B $\rightarrow$ L405B} & {0.195} & {\textbf{0.058}} & {0.137} & {0.201} & {\textbf{0.150}} & {0.051} & {0.252} & {\textbf{0.181}} & {0.071} & {0.095} & {\textbf{0.041}} & {0.054} & {0.340} & {\textbf{0.298}} & {0.042} & {0.488} & {\textbf{0.169}} & {0.319} & {0.262} & {\textbf{0.150}} & {0.112} \\
\textbf{L3B $\rightarrow$ L8B} & {0.119} & {\textbf{0.089}} & {0.031} & {0.410} & {\textbf{0.395}} & {0.015} & {0.387} & {\textbf{0.381}} & {0.006} & {0.196} & {\textbf{0.185}} & {0.011} & {0.485} & {\textbf{0.473}} & {0.012} & {0.522} & {\textbf{0.501}} & {0.020} & {0.353} & {\textbf{0.337}} & {0.016} \\
\textbf{L3B $\rightarrow$ L70B} & {0.058} & {\textbf{0.029}} & {0.028} & {\textbf{0.200}} & {0.208} & {-0.008} & {0.235} & {\textbf{0.235}} & {0.000} & {0.084} & {\textbf{0.079}} & {0.005} & {0.403} & {\textbf{0.392}} & {0.011} & {0.267} & {\textbf{0.164}} & {0.103} & {0.208} & {\textbf{0.184}} & {0.023} \\
\textbf{L3B $\rightarrow$ L405B} & {0.122} & {\textbf{0.016}} & {0.106} & {0.180} & {\textbf{0.109}} & {0.071} & {0.251} & {\textbf{0.174}} & {0.076} & {0.073} & {\textbf{0.060}} & {0.014} & {0.345} & {\textbf{0.290}} & {0.055} & {0.483} & {\textbf{0.200}} & {0.283} & {0.242} & {\textbf{0.142}} & {0.101} \\
\textbf{L8B $\rightarrow$ L70B} & {\textbf{0.021}} & {0.024} & {-0.002} & {\textbf{0.186}} & {0.209} & {-0.023} & {\textbf{0.232}} & {0.242} & {-0.011} & {0.064} & {\textbf{0.018}} & {0.047} & {0.410} & {\textbf{0.399}} & {0.010} & {0.256} & {\textbf{0.159}} & {0.097} & {0.195} & {\textbf{0.175}} & {0.020} \\
\textbf{L8B $\rightarrow$ L405B} & {0.050} & {\textbf{0.019}} & {0.030} & {0.166} & {\textbf{0.112}} & {0.054} & {0.239} & {\textbf{0.190}} & {0.049} & {0.038} & {\textbf{0.021}} & {0.017} & {0.348} & {\textbf{0.270}} & {0.078} & {0.379} & {\textbf{0.124}} & {0.254} & {0.203} & {\textbf{0.123}} & {0.081} \\
\textbf{L70B $\rightarrow$ L405B} & {\textbf{0.008}} & {0.011} & {-0.002} & {0.122} & {\textbf{0.097}} & {0.025} & {0.192} & {\textbf{0.189}} & {0.003} & {\textbf{0.022}} & {0.042} & {-0.021} & {0.325} & {\textbf{0.302}} & {0.023} & {0.152} & {\textbf{0.048}} & {0.105} & {0.137} & {\textbf{0.115}} & {0.022} \\
\textbf{4o-Mini $\rightarrow$ Q32Bc} & {0.048} & {\textbf{0.003}} & {0.045} & {0.280} & {\textbf{0.278}} & {0.002} & {0.158} & {\textbf{0.157}} & {0.000} & {0.077} & {\textbf{0.071}} & {0.006} & {0.294} & {\textbf{0.273}} & {0.021} & {0.044} & {\textbf{0.019}} & {0.025} & {0.150} & {\textbf{0.134}} & {0.017} \\
\textbf{4o-Mini $\rightarrow$ Q72B} & {0.026} & {\textbf{0.012}} & {0.014} & {0.216} & {\textbf{0.214}} & {0.002} & {0.107} & {\textbf{0.105}} & {0.002} & {\textbf{0.067}} & {0.070} & {-0.003} & {0.294} & {\textbf{0.288}} & {0.007} & {0.014} & {\textbf{0.004}} & {0.010} & {0.121} & {\textbf{0.115}} & {0.005} \\
\textbf{4o-Mini $\rightarrow$ 4o} & {0.070} & {\textbf{0.005}} & {0.065} & {0.201} & {\textbf{0.179}} & {0.022} & {0.182} & {\textbf{0.156}} & {0.025} & {0.052} & {\textbf{0.039}} & {0.013} & {0.235} & {\textbf{0.203}} & {0.032} & {0.032} & {\textbf{0.001}} & {0.031} & {0.128} & {\textbf{0.097}} & {0.031} \\
\textbf{Q32Bc $\rightarrow$ Q72B} & {0.035} & {\textbf{0.000}} & {0.034} & {0.270} & {\textbf{0.258}} & {0.011} & {\textbf{0.153}} & {0.154} & {-0.002} & {0.145} & {\textbf{0.126}} & {0.019} & {0.365} & {\textbf{0.357}} & {0.008} & {0.097} & {\textbf{0.039}} & {0.059} & {0.177} & {\textbf{0.156}} & {0.022} \\
\textbf{Q32Bc $\rightarrow$ 4o} & {0.001} & {\textbf{0.000}} & {0.001} & {0.259} & {\textbf{0.203}} & {0.056} & {0.211} & {\textbf{0.128}} & {0.082} & {0.110} & {\textbf{0.069}} & {0.042} & {0.295} & {\textbf{0.245}} & {0.050} & {0.238} & {\textbf{0.007}} & {0.231} & {0.186} & {\textbf{0.109}} & {0.077} \\
\textbf{Q72B $\rightarrow$ 4o} & {0.027} & {\textbf{0.004}} & {0.023} & {0.254} & {\textbf{0.205}} & {0.049} & {0.148} & {\textbf{0.027}} & {0.121} & {0.057} & {\textbf{0.049}} & {0.008} & {0.293} & {\textbf{0.256}} & {0.037} & {0.040} & {\textbf{0.000}} & {0.040} & {0.136} & {\textbf{0.090}} & {0.046} \\
\midrule
\textbf{Average} & {0.070} & {\textbf{0.033}} & {0.037} & {0.248} & {\textbf{0.227}} & {0.022} & {0.238} & {\textbf{0.208}} & {0.030} & {0.113} & {\textbf{0.092}} & {0.020} & {0.369} & {\textbf{0.344}} & {0.025} & {0.288} & {\textbf{0.178}} & {0.110} & {0.221} & {\textbf{0.181}} & {0.041} \\
\bottomrule
\end{tabular}
\end{adjustbox}
\end{table}

\end{document}